\documentclass{sigchi-ext}
\usepackage[T1]{fontenc}
\usepackage{textcomp}
\usepackage[scaled=.92]{helvet} 
\usepackage{graphicx} 
\usepackage{balance}  
\usepackage{makecell}
\usepackage{multirow}

\usepackage[utf8]{inputenc} 

\newif\ifANONYMOUS

\usepackage{todonotes}
\usepackage{changes}
\makeatletter
\newsavebox{\@temp@sb}
\newcommand{\deftodo}[4]{%
    \expandafter\def\csname todo#1\endcsname##1{%
        \sbox{\@temp@sb}{\parbox{\linewidth-8pt}{#2: {##1}}}%
        \todo[inline,color=#3]{\textcolor{#4}{\usebox{\@temp@sb}}}}
}
\makeatother
\deftodo{sg}{St\'ephane}{cyan}{black}
\deftodo{ym}{Yazan}{orange}{black}
\deftodo{an}{Amro}{yellow}{black}
\deftodo{tk}{Tim}{green}{black}
\deftodo{igor}{Igor}{magenta}{black}
\deftodo{cn}{Christophe}{purple}{white}

\def\plaintitle{Towards Explainability for a Civilian UAV Fleet Management using an Agent-based Approach} 
\ifANONYMOUS
    \def\plainauthor{}
    \def\plainauthors{}
\else
    \def\plainauthor{Yazan Mualla}
    \def\plainauthors{\plainauthor\ et al.}
\fi
\def\plainkeywords{Intelligent Transport Systems, Explainability, Multiagent Systems} 
\definecolor{linkColor}{RGB}{6,125,233}

\title{\plaintitle}

\numberofauthors{6}
\author{%
    \ifANONYMOUS\else%
        \alignauthor{%
            \textbf{Yazan Mualla} \\
                \affaddr{CIAD, Univ. Bourgogne Franche-Comt\'e, UTBM, \\90010 Belfort, France} \\
                \email{yazan.mualla@utbm.fr} \\[\baselineskip]
            \textbf{Amro Najjar} \\
                \affaddr{AI-Robolab/ICR, Computer Science and Communications, University of Luxembourg, \\ 4365 Esch-sur-Alzette, Luxembourg} \\[\baselineskip]
            \textbf{Timotheus Kampik} \\
                \affaddr{Department of Computing Science, Ume\.a University, \\ 90736 Ume\.a, Sweden} \\[\baselineskip]
             \textbf{Igor Haman Tchappi} \\
                \affaddr{University of Ngaoundere \\ 454 Ngaoundere, Cameroon}
        }
        \alignauthor{%
            \textbf{St\'ephane Galland} \\
                \affaddr{CIAD, Univ. Bourgogne Franche-Comt\'e, UTBM, \\ 90010 Belfort, France} \\[\baselineskip]
            \textbf{Christophe Nicolle} \\
                \affaddr{CIAD, Univ. Bourgogne Franche-Comt\'e, UB, \\ 21000 Dijon, France}
        }
    \fi
}

\hypersetup{%
  pdftitle={\plaintitle},
  pdfauthor={\plainauthors},
  pdfkeywords={\plainkeywords},
  bookmarksnumbered,
  pdfstartview={FitH},
  colorlinks,
  citecolor=black,
  filecolor=black,
  linkcolor=black,
  urlcolor=linkColor,
  breaklinks=true,
}

\newcounter{blindedreference}
\newcommand{\selfcite}[1]{%
    \ifANONYMOUS%
        \expandafter\ifx\csname blindedreference#1\endcsname\relax%
            \stepcounter{blindedreference}%
            \expandafter\xdef\csname     blindedreference#1\endcsname{\theblindedreference}%
        \fi%
        {[\textbf{BLINDED REFERENCE \csname blindedreference#1\endcsname}]}%
    \else%
        \cite{#1}%
    \fi%
}

\begin{document}
\justify
\CopyrightYear{2019}
\setcopyright{rightsretained}
\conferenceinfo{Auto UI}{'19 Utrecht, Netherlands}
\isbn{isbn}
\doi{doi}
\copyrightinfo{\acmcopyright}

\maketitle


\begin{abstract}
This paper presents an initial design concept and specification of a civilian Unmanned Aerial Vehicle (UAV) management simulation system that focuses on explainability for the human-in-the-loop control of semi-autonomous UAVs. The goal of the system is to facilitate the operator intervention in critical scenarios (e.g. avoid safety issues or financial risks). Explainability is supported via user-friendly abstractions on Belief-Desire-Intention agents. To evaluate the effectiveness of the system, a human-computer interaction study is proposed.
\end{abstract}

\keywords{\plainkeywords}



\section{Introduction}

With the rapid increase of the world's urban population, the infrastructure of the constantly expanding metropolitan areas is subject to immense pressure. To meet the growing demand for sustainable urban environments and improve the quality of life for citizens, municipalities will increasingly rely on novel transport solutions. In particular, Unmanned Aerial Vehicles (UAVs), commonly known as drones, are expected to have a crucial role in future smart cities thanks to relevant features such as autonomy, flexibility, mobility, adaptive altitude, and small dimensions~\cite{mualla2019between}. Therefore, over the past few years, an increasing number of public and private research laboratories have been working on civilian, small, and human‐friendly drones.

Still, several concerns exist regarding the possible consequences of introducing UAVs in crowded urban areas, especially regarding people's safety, e.g. if a mechanical failure causes a crash. To guarantee it is safe for UAVs to fly close to human crowds and to reduce costs, different scenarios must be modeled and tested. Yet, regulations restrict the use of UAVs in cities. Additionally, to perform tests with real UAVs, one needs access to expensive hardware. Moreover, field tests usually consume a considerable amount of time and require trained and skilled people to pilot and maintain the UAVs. Furthermore, on the field, it may also be hard to reproduce the same scenario several times~\cite{lorig2015measuring}. In this context, the development of simulation frameworks that allow transferring real world scenarios into executable models using computer simulation frameworks, i.e. simulating UAVs activities in a digital environment are highly relevant~\cite{mualla2019agent},~\cite{mualla2018comparison}. However, simulation frameworks have their own drawbacks, i.e. it is impossible to fully reproduce the real environment.

Intelligent software agents have been established as a suitable technique for implementing autonomous control and decision making in computer systems~\cite{wooldridge1995intelligent}. It is also used for different simulation applications in general \cite{najjar2017aquaman, mualla2018agentoil, najjar2017aquamanWI} and for semi-autonomous systems in different domains \cite{muallaOilCPS19} in particular. 
The use of Agent-Based Simulation (ABS) frameworks in UAVs is gaining more interest in complex civilian application scenarios where coordination and cooperation are necessary~\cite{mualla2018comparison}. ABS models a set of interacting intelligent entities that reflect, within an artificial environment, the relationships in the real world~\cite{wooldridge1995intelligent}. Due to operational costs, safety concerns, and legal regulations, ABS is commonly used to implement models and conduct tests. This has resulted in a range of research works addressing ABS in UAVs~\cite{mualla2019agent}. The results make ABS a natural step forward towards better understanding and managing the complexity of today's business and social systems.


However, as UAVs are like any other robot, communication with humans is a challenge, since the human user is not by default capable of understanding the robot's State of Mind (SoM)~\cite{hellstrom2018understandable}.
This problem is even more accentuate in the case of UAV since--as has been confirmed by recent studies in the literature~\cite{bainbridge2008effect, hastie2017trust}--remote robots tend to instill less trust than those co-located. For this reason, working with remote robots is  more challenging task specially in high-stakes scenarios such as flying UAVs in the urban environment. To overcome this challenge, this paper relies on the recent advances of the domain of  eXplainable Artificial Intelligence (XAI) \cite{guidotti2019survey}, \cite{XAISLR}, \cite{calvaresi2019XAI} in order to trace the decisions of the agents, and enable the validation of their behaviors when they are applied in a fleet of civilian UAVs that are interacting with other objects in the air or in the smart city.  
In this paper, we present a conceptual design of a MultiAgent System (MAS) that simulate the civilian UAVs' fleet, and provides tools for building the explainability of the system.

The rest of this paper is structured as follows. 
Firstly an overview of related works is given.
Then, the explainable MAS for aerial transportation is highlighted and a use case in a smart city is presented.
Finally, a conclusion and future research directions are presented.


\section{Related Work}\label{sec-related-works}

\subsection{The rise of XAI}
In the last couple of years, work on XAI is gaining momentum both in research and industry. Primarily, this surge is explained by the success of black-box machine learning mechanisms whose inner workings are incomprehensible by human users~\cite{gunning2017explainable}. Therefore, XAI aims to ''open`` the black-box and explain the sometimes intriguing results of its mechanisms e.g. a Deep Neural Network (DNN) mistakenly classifying a tomato as a dog~\cite{szegedy2013intriguing}. In contrast to this data-driven explainability~\cite{XAISLR}, more recently, this tendency has been extended to explain the complex behavior of goal-driven systems such as robots/agents~\cite{XAISLR}\cite{hellstrom2018understandable} since: \emph{(i)} as has been shown in the literature, humans tend to assume that these agents \& robots have their own SoM~\cite{hellstrom2018understandable} and that with the absence of a proper explanation, the user will come up with an explanation that might be flawed or erroneous, \emph{(ii)} these agents/robots are expected to be omnipresent in the daily lives of their users (e.g. social assistive robots and virtual assistants). 

\subsection{Explainability of UAVs}
Recently, both data-driven and goal-driven explainability have been introduced in UAVs. In the former case, XAI aims to \emph{interpret} the opaque machine learning mechanisms used by those UAV to analyze the input originating from their rich data streams.   One ongoing work investigates how to interpret the decisions of convolution neural networks analyzing aviation related images as inputs~\cite{dolph2018towards}. Explainability is explored by reviewing three feature visualization methods in a layer-by-layer approach. 
In the latter case, explainability aims to make the autonomous behavior of the UAV understandable. One work is based on fuzzy logic: the explainable model is presented on a visual platform in the format of if-then rules derived from the fuzzy inference model~\cite{keneni2019evolving}. 
Our approach belongs to the goal-driven case and is different than other related work as it relies on a decentralized solution using MAS. This choice is supported by the fact that the management of a UAV fleet must consider the physical distance between UAVs and other actors in the system. Additionally, autonomous agents represent in our opinion an adequate implementation of the autonomy of UAVs. The choice of Belief-Desire-Intention (BDI)~\cite{bratman1987intention} model is to support the explainability in the UAV fleet as detailed in the next section. 

\section{Explainable MAS for Aerial Transportation}\label{sec-explainable-xai}

Agent architectures, like BDI model, are frequently applied to equip UAVs with greater autonomy.
By designing proactive agents that control UAVs, the latter become capable of autonomously managing their actions and behavior to reach their goals~\cite{arokiasami2016interoperable}.
As shown by Padgham et al.~\cite{Padgham2011IntegratingBR}, BDI models can facilitate explainable agency, as they provide a well-structured snapshot of agent internals at any point of time.
The BDI paradigm is frequently applied in ABS. 
Adam and Gaudou~\cite{adam2016bdi} present an extensive analysis and evaluation of approaches for integrating BDI models in ABS and highlight the previously mentioned benefits of BDI models as a way to implement descriptive agents that use richer and--from a human perspective-- better interpretable abstractions than purely reactive agents.



\begin{figure}[t]
  \includegraphics[width=1\linewidth]{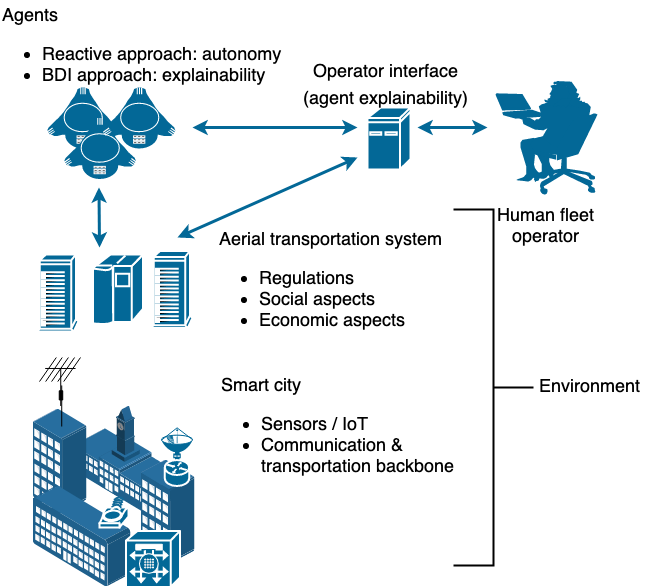} 
  \caption{Contextual Model}
  \label{fig:Contextual_Model}
\end{figure}

Figure~\ref{fig:Contextual_Model} outlines the contextual model of our work; in the bottom left, we see the technical requirements and infrastructure in a smart city, into which the aerial transportation management system (middle left) is embedded. In the top, the reactive agents guarantee autonomy when performing the normal tasks of an intelligent transport system,  while there is a need for explainability that is associated with BDI agents and allows for human-in-the-loop control (top right). The UAVs in the fleet explain to the human their autonomous behavior and decisions along with any deviation from the planned mission. 

%
\begin{table}
 \footnotesize
     \def\arraystretch{1.5}
     \setlength\tabcolsep{10pt}
    \centering
    \caption{A high-level perspective of an agent's BDI snapshot.}
    \begin{tabular}{ |l|c| } 
     \hline
     \multirow{5}{5em}{\textbf{Beliefs}} & \texttt{package\_1: \{ to: Townhouse\_27a\}} \\ 
      & \texttt{package\_2: \{ to: School\_1 \}}  \\ 
      & \makecell{\texttt{Townhouse\_27a:} \\ \texttt{\{ time: 20min, charger: no \}}}  \\
      & \makecell{\texttt{School\_1:} \\ \texttt{\{ time: 20min, charger: yes \}}}  \\
      & \texttt{batteryTime: 22min}  \\
     \hline
     \multirow{2}{2em}{\textbf{Desires}} & \texttt{deliver: package\_1} \\
      & \texttt{deliver: package\_2}  \\ 
     \hline
     \textbf{Intentions} & \texttt{deliver: package\_2} \\
     \hline
     \textbf{Plans} & \texttt{moveTo(School\_1)}  \\ 
     \hline
    \end{tabular}
    \label{table:snapshot}
\end{table}
%
Table~\ref{table:snapshot} shows a potential example of a high-level perspective of an agent's BDI snapshot. 


\section{DroneAgent-Delivery: a use case of human or package transportation in a smart city}\label{sec-UAV-agent-delivery}



The use case is about investigating the role of XAI in the communication between drones and humans in the context of human or package transportation in a smart city. 

In the scenario, one operator is in charge of several drones that will provide  transportation services to clients. These drones will autonomously conduct tasks and take decisions when needed. Additionally  they need to communicate and discuss with each other and may cooperate to complete a specific task. The drones will explain to the Operator Assistant Agent (OAA) the progress of the mission including the unexpected events along with the decisions made by them. 


\begin{figure}[t]
  \includegraphics[width=1\linewidth]{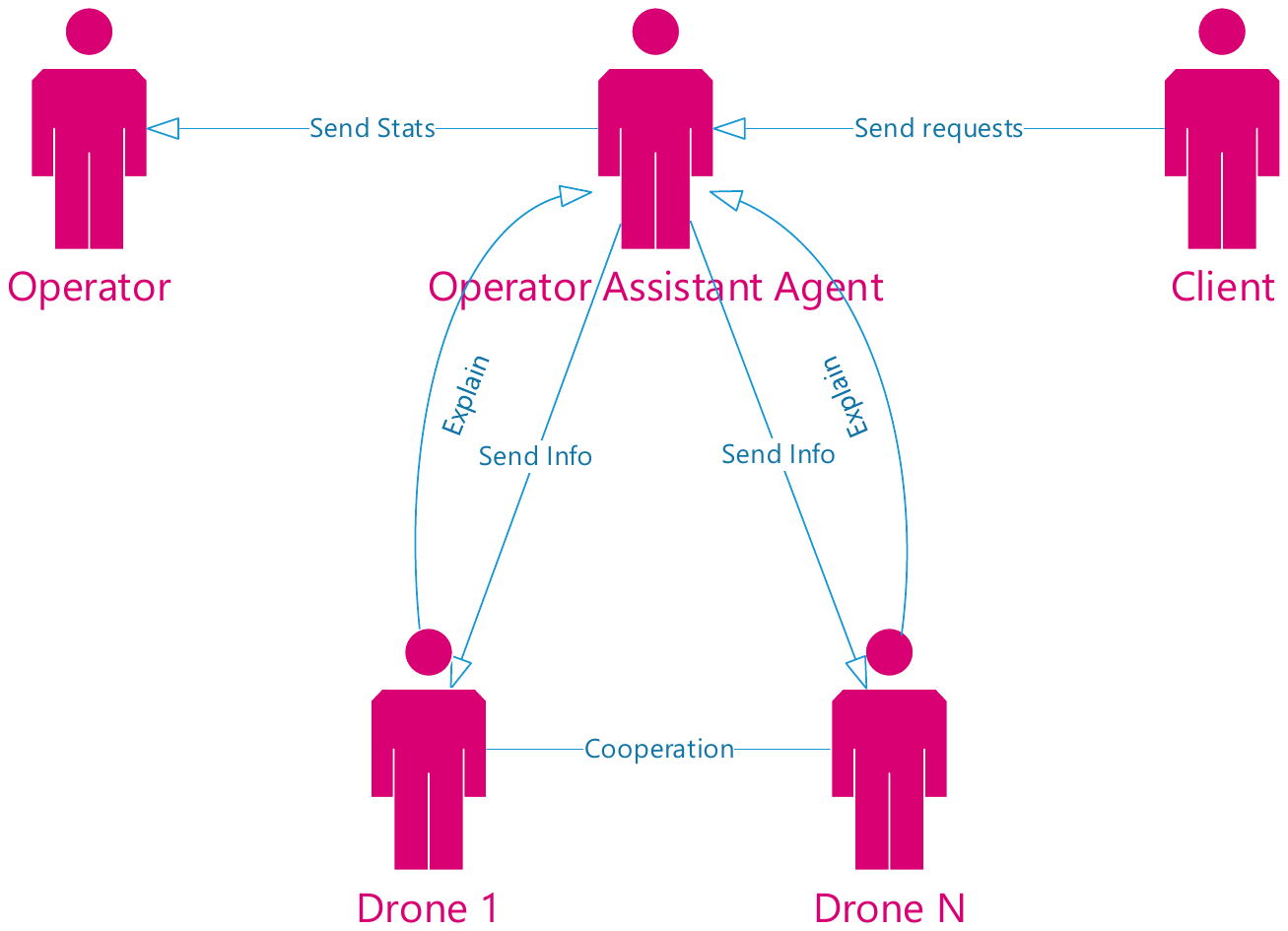}
  \caption{Interaction of Actors in the System}
  \label{fig:Actors}
\end{figure}

Figure \ref{fig:Actors} shows the interaction between the actors in the proposed use case. 
In the following, the steps of the use case are detailed:
\begin{enumerate}
    \item When a client puts a request for transporting a package/passenger, a notification is sent to the OAA.
\item The OAA will send it to all drones, so all drones are connected with each other and with the OAA using an assumed reliable network.

\item Drones that are near, with a specific radius, to the package/passenger will coordinate to complete the transportation mission. 
The decentralized coordination (without the approval of the operator) can be for several reasons:
\begin{itemize}
    \item Who will deliver the package/passenger according to constraints: actual distance to the package/passenger, battery
size, having other packages/passengers in hand, having a mission with a near destination, etc.

    \item The load is heavy, and it needs several drones working as a swarm to lift it.

    \item There is a need to cooperate to deliver the package/passenger between several drones, where each drone delivers the load part of the way and then hands it to another drone.
\end{itemize}
    
\item Every drone will explain its own arguments to the OAA. At the end, the result of the coordination discussion will be sent to the OAA that will show it with/without filtration to the operator.

\item If the package/passenger is picked up by a drone from a competitor (external events), we have two situations:
\begin{itemize}
    \item The client sends to the OAA that the package or passenger is picked up, and the assistant
agent will inform the assigned drone to stop the mission;
\item The client does not send a notification that the package/passenger is picked up (because of selfishness or
laziness). In this situation, the drone will go to the place and observe the absence of the package/passenger, and it needs to explain
this to the OAA.

\end{itemize}

\item The explanation needed from the drone is generally about the mission progress, its decisions and its status, e.g. the drone needs charging and that is why it ignores a nearby package/passenger, etc. Other kinds of explanation needed from the drones are the unexpected events, e.g. the drone arrives at the package location and see that is it damaged, or not according to the description (maybe heavier).
\item the OAA may filter the explanations received from the drones to give a summary of the most important explanations to avoid overwhelming the operator with a lot of details.
\item The operator at any time can either look at full explanations from all the drones or only the results filtered by the OAA.

\end{enumerate}



\subsection{Evaluation of the autonomy and explainability}

For evaluation, there is a need for an ABS to simulate an application of drones’ autonomy
and explainability.
The evaluation is performed as a human-computer interaction study.
The participants will try the simulation and fill out a  questionnaire to discover if a human user (as the operator) can understand
the explanations provided by the UAVs. The questionnaire will be built specifically according to the XAI metrics provided in the literature~\cite{hoffman2018metrics}.

The evaluation will have three levels:
\begin{itemize}
    \item First level: we test the explainability aspect. The experiment requires dividing the participants into three groups. The first, second and third groups will try the simulation without any explanation, with full explanation, and with filtered explanation, respectively.  
\item Second level: we test the shared autonomy aspect. We ask the participants to perform a tedious task like filling
an excel sheet while the simulation is running in two modes (First mode: The operator only looks at the full
explanations by all drones. Second mode: The operator is looking at only the filtered explanations from the OAA). Then we measure in what mode the participants completed more percentage of the tedious task
assigned to them.
\item Third level: we get the results and statistics from the simulation, e.g. how many packages delivered, if a drone
did not deliver any package, the distance moved by all drones, etc. The operator benefits from these results to change the initial parameters of the simulation for the
second run and check if the results are better. 

\end{itemize}

\section{Conclusion and future work}\label{sec-conclusion}
This paper presented a concept and a specification for an agent-based civilian UAV fleet management approach with a focus on explainability.
The presented work is an initial step towards the goal of providing agent-based tools that allow for a human-in-the-loop approach supporting semi-autonomous UAV fleet management.
In particular, the following work is of importance:
\begin{description}
    \item[Engineering research] should be conducted to design an explainable multiagent management system whose architecture is sufficiently generic to be applied for use cases beyond UAV fleet management. 
    \item[Empirical assessment] should evaluate the effectiveness of the approach in human-computer interaction studies.
\end{description}

\section{Acknowledgements}

\small
This work is supported by the Regional Council of Bourgogne Franche-Comté (RBFC, France) within the project UrbanFly 20174-06234/06242. The first author thanks \emph{Cedric Paquet} for his remarks regarding the evaluation.

\justify
\bibliographystyle{SIGCHI-Reference-Format}
\bibliography{sample}



\end{document}